\documentclass[runningheads]{llncs}
\usepackage{booktabs}
\usepackage{amsmath,amssymb,amsfonts}
\usepackage{mathrsfs}
\usepackage[T1]{fontenc}
\usepackage{graphicx}
\begin{document}
\title{Implicit Bias in Deep Linear Discriminant Analysis}

\author{Jiawen Li\inst{1}\orcidID{0009-0002-1449-2803}}
\authorrunning{J. Li} 

\institute{School of Computer Science and Engineering, University of New South Wales, \\ 
Kensington, Sydney 2052, New South Wales, Australia \\
\email{jiawen.li12@student.unsw.edu.au}}

\maketitle 

\begin{abstract}
While the Implicit Bias(or Implicit Regularization) of standard loss functions has been studied, the optimization geometry induced by discriminative metric-learning objectives remains largely unexplored. To the best of our knowledge, this paper presents an initial theoretical analysis of the implicit regularization induced by the Deep LDA—a scale-invariant objective designed to minimize intra-class variance and maximize inter-class distance. By analyzing the gradient flow of the loss on a L-layer diagonal linear network, we prove that under balanced initialization, the network architecture transforms standard additive gradient updates into multiplicative weight updates, which demonstrates an automatic conservation of the $||\cdot||_{2/L}$ quasi-norm.

\end{abstract}

\section{Introduction}\label{sec1}
Deep learning has been successfully applied in numerous domains. In the realm of deep learning, researchers widely believed there’s an implicit factor that acts as a constraint in the training process of gradient-based models, ensuring their weights do not explode, which is referred to as “Implicit Bias”.
With the involvement of more real-time applications, implicit bias has attracted increasing attention in recent years. However, these theoretical breakthroughs are still majorly confined to loss functions with exponential tails (eg, Cross-Entropy) or square losses.

In classical statistical learning, Fisher Discriminant Analysis(FDA) is a classical method was proposed by R.A. Fisher in 1936, aiming to preserve information of data after reducing dimensions in a classification task \cite{FDA}. In Matthias Dorfer's group’s works around 2015 , they propose a new loss function based on FDA, Deep Linear Discriminant Analysis(Deep LDA) \cite{DeepLDA}, with the same design concept, to maximize the inter-class distance while minimizing the intra one. While empirical studies demonstrate that such discriminative objectives yield highly separable features, their Implicit Bias still remains largely unexplored.

\section{Literature Review}\label{sec2}
\begin{figure}[h]
\centering
\includegraphics[width=1\linewidth]{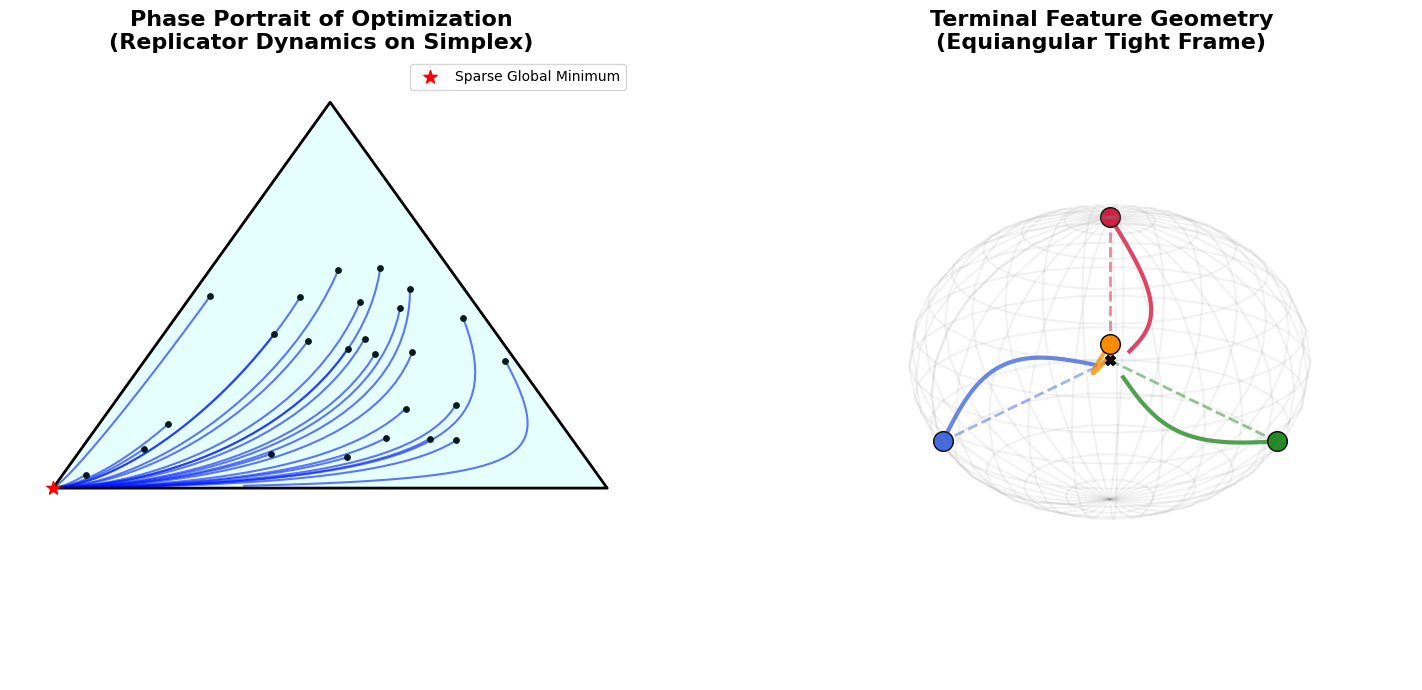}
\caption{The Optimization in Simplex and the Implicit Bias}
\label{fig:1}
\end{figure}

Implicit Bias, as previously mentioned, is one of the main hypothesis try to explain models’  good generalization performance, assuming there exists an implicit inclination of the model that select features. Understanding them better could lead more mathematical rigorous design to the loss function, and further imply the future improvement of model structures. As the left subplot show in figure.\ref{fig:1}, the gradient decent have an inclination that lead a same solution. Instead of using the traditional method to direct discuss its bound and convergence, the Implicit Bias provides another framework to analyze the reasons causing it in terms of the dynamics between features that tend to be bound by a constraint. Just as the right subplot illustrates,Equiangular Tight Frame (ETF) are a phenomenon that vectors that are distribute even with identical angles between each vectors \cite{collapse}, this feature collapse is explicitly design in DeepLDA formulas, the deeply learned features repel each other and unfold from the origin,which makes a regular tetrahedron. 

As for discussing the implicit regularization in logistic functions, the most famous research is Daniel Soudry’s work in 2018, finding that the weight vector is approaching infinity while still approximating the direction of $||\cdot||_{2}$ norm \cite{Cross1}. This result explains why the generalization performance still improves even after the loss function becomes stable in classification tasks. After a while, the conclusion from Ziwei Ji and Matus Telgarsky discover for logistic-type loss functions\cite{selfAlign}, the gradient descent would lead to self-alignment between the weights of each layer. The square loss is the most common type of loss function, hence it gets analyzed earlier. Suriya Gunasekar’s group focuses on matrix decomposition,and discovered that certain initialization could lead model weights to low-rank\cite{Square1}. Just in next year, the same Suriya Gunasekar find the square loss in CNN with a stride would trigger sparsity in domain frequency\cite{Square2}.

Albeit the previous works talk about all prove to be have an Implicit Bias, but not loss function could prove to have an explicit mathematical Implicit Bias as Noam Razin suggested \cite{Unexplain}. Building on Razin’s observation, it becomes imperative to explore whether a fundamentally different class of objectives could reveal entirely new optimization geometries. For instance, Deep LDA, as a famous loss adopting in wide range of applications especially in Computer Vision(for example, Deep LDA use in COVID-19 diagnose\cite{LDA_USE}), but the Implicit Bias of it still remains as an open question. To fill the gap, our work formulates the gradient flow of the Rayleigh Quotient, revealing that this specific metric-learning objective explicitly imposes a strict quasi-norm conservation.

\section{Methodology}\label{sec3}
We further extend this implicit bias analysis to a scale-invariant situation and try to figure out why and how Deep LDA induces feature sparsity. When knowing the specific Rayleigh Quotient objective, the gradient flow could be further simplified and reveal the whole optimization process by using continuous-time dynamics, in that way, if the reduction of redundant weights is happening with the more severe multiplicative penalty that network depth brought, exact quasi-norm conservation could get work.

Diagonal Linear Networks(DLNs) is a standard proxy model to discuss the model regularization \cite{DLNs}, due to its special property allowing analysis focus on depth of network without other distributions. Hence for our theoretical setup, we adopt an $L$-layer DLN to rigorously analyze the gradient flow dynamics of the Deep LDA objective. In this paper, we only analyze the continuous gradient flow under balanced initialization, not including non-linear activation functions or other complex network architectures.

Although Lyu and Li \cite{approachDirection} proved that gradient descent on homogeneous networks with exponential-tailed losses converges in direction to a maximum-margin solution, but these results rely on positive-degree homogeneity, whereas the Deep LDA objective is homogeneous of degree zero, which ushers to a fundamentally different geometric constraint as we show in this work.

\subsection{Theoretical Assumptions}
To simplify the conditions of the analysis while still keeping the effects of the depth of a neural network, we make a few setups and assumptions for our basic methodology:

1.\textbf{Balance Initialization}: It assumes the change rate of each layers in equivalent, this helping us convert the weights of the network into a form of norm to analyze, ensuring symmetry across layers.

2.\textbf{Continuous Gradient Flow}: For most traditional neural networks, analyzing the continuous-time ordinary differential equations (ODEs) rather than discrete update steps is a standard and reasonable mathematical proxy to understand optimization\cite{flow1}\cite{flow2}. This also indicates our derivation still lacks verification for discrete models(for example, like Hopfield Network\cite{Hopfield}).

3. \textbf{Diagonal Linear Architecture}: Because each feature dimension is processed independently without cross-dimensional interactions, helping the framework theory only discuss the effect brought by depth while eliminating interactions from complex impacts brought by correlations.

To conclude, the methodology employs balance initialization, continuous gradient flow, and a diagonal linear architecture to isolate the effects of network depth. These simplifications enable symmetric layer analysis while restricting the framework to continuous, non-discrete models.

\subsection{Diagonal Linear Networks(DLNs)}
First, we defined a deep linear network with $L$ layers representing the mapping from input to output. $u^{(1)}$ represent the weights of the first layer, and $u^{(L)}$ is the last one. In addition, all of them are restricted to diagonal matrices. Those sequences of weights are not fully connected in the definition, which means we are not just simply defining dense matrices but isolated parallel paths for each feature dimension.

\begin{equation}
w_i = \prod_{k=1}^L u_i^{(k)}
\label{eq:1}
\end{equation}

\subsubsection{Conservation Lemma for DLNs}
We need to prove there exist a conservation for DLNs first, for a better overview and to reduce the computation for Deep LDA in the next subsection. For any ${k}^{th}$ layer, we could get the effects each element in this layer if it gets impacted by the weight $w$ using the chain rule:
\begin{equation}
\frac{\partial \mathcal{L}}{\partial u_i^{(k)}} = \frac{\partial \mathcal{L}}{\partial w_i} \frac{\partial w_i}{\partial u_i^{(k)}} = \frac{\partial \mathcal{L}}{\partial w_i} \left( \prod_{m \neq k} u_i^{(m)} \right) = \frac{\partial \mathcal{L}}{\partial w_i} \frac{w_i}{u_i^{(k)}}
\label{eq:2}
\end{equation}

Assume variable $t$ is the current step, for standard gradient descent, the expression for the element $u_k$ would be like:
\begin{equation}
u_i^{(k)}(t+1) = u_i^{(k)}(t) - \eta \frac{\partial \mathcal{L}}{\partial u_i^{(k)}}
\label{eq:3}
\end{equation}

In eq.\eqref{eq:3}, we transform it into a form easy to become derivative about $t$. First move $u_i^{(k)}(t)$ to the left, and then divide $\eta$ in same time.
\begin{equation}
\frac{u_i^{(k)}(t+1) + u_i^{(k)}(t)}{\eta } = - \frac{\partial \mathcal{L}}{\partial u_i^{(k)}}
\label{eq:4}
\end{equation}

For getting the analytical solution, we take the limit of the eq.\eqref{eq:4},and assume $\eta \rightarrow 0$. Taking the continuous-time limit $\eta \rightarrow 0$,eq.\eqref{eq:4} becomes the gradient flow ODE in eq.\eqref{eq:5}.
\begin{equation}
\lim_{\eta \to 0} \frac{u_i^{(k)}(t+\eta) - u_i^{(k)}(t)}{\eta} = \frac{du_i^{(k)}}{dt} = - \frac{\partial \mathcal{L}}{\partial u_i^{(k)}}
\label{eq:5}
\end{equation}

Substitute $\frac{\partial \mathcal{L}}{\partial u_i^{(k)}}$  in eq.\eqref{eq:2} as $\frac{du_i^{(k)}}{dt}$,  we could get eq.\eqref{eq:6}.
\begin{equation}
\frac{du_i^{(k)}}{dt} = - \left( \frac{\partial \mathcal{L}}{\partial w_i} \frac{w_i}{u_i^{(k)}} \right)
\label{eq:6}
\end{equation}

Next step, we multiply the right-hand side(RHS) and the left-hand side(LHS) at the same time to move the denominator to the left:
\begin{equation}
u_i^{(k)} \frac{du_i^{(k)}}{dt} = - w_i \frac{\partial \mathcal{L}}{\partial w_i}
\label{eq:7}
\end{equation}

Due to the basic calculus rule $\frac{1}{2} \frac{d}{dt}(x^2) = x \frac{dx}{dt}$, we could further simplify it as form in eq.\eqref{eq:8}.

\begin{equation}
\frac{1}{2} \frac{d}{dt} \left( u_i^{(k)} \right)^2 = - w_i \frac{\partial \mathcal{L}}{\partial w_i}
\label{eq:8}
\end{equation}

By observing the eq.\eqref{eq:8}, we could see whether what $k^{th}$ layer is, the right hand side is kept constant. Hence we could get $\frac{d}{dt} \left( u_i^{(k)} \right)^2 = \frac{d}{dt} \left( u_i^{(m)} \right)^2$.

And take the integral to both sides:
\begin{equation}
\left( u_i^{(k)}(t) \right)^2 - \left( u_i^{(m)}(t) \right)^2 = C
\label{eq:9}
\end{equation}

Furthermore, we utilize balanced initialization, assume $u_i^{(1)}(0) = u_i^{(2)}(0) = \dots = u_i^{(L)}(0)$,hence for $t=0$, there always be $\left( u_i^{(k)}(t) \right)^2 = \left( u_i^{(m)}(t) \right)^2$. Additionally, the weight is always positive, so we could further conclude:

\begin{equation}
u_i^{(1)}(t) = u_i^{(2)}(t) = \dots = u_i^{(L)}(t)
\label{eq:10}
\end{equation}

As we defined in eq.\eqref{eq:1}, the form of $w_i(t)$ could write as :
\begin{equation}
\ w_i(t) = \left( u_i^{(k)}(t) \right)^L
,\
u_i^{(k)}(t) = w_i(t)^{\frac{1}{L}}
\label{eq:11}
\end{equation}

\subsection{Implicit Bias in Deep LDA}
\begin{equation}
\mathcal{L}(w) = \frac{w^\top S_w w}{w^\top S_b w}
\label{eq:12}
\end{equation}

The eq.\eqref{eq:12} shows the main concept of this scale-invariant objective, it assumes the loss of the whole dataset could be formulated as formed in a Rayleigh Quotient. We express the loss as the ratio of the intra-class variance and the inter-class variance, as Deep LDA(or Generalized Rayleigh Quotient) is defined, when $w \neq 0$ and $w^\top B w \neq 0$.

When we multiply the $w$ by the factor $a$, the loss function of Deep LDA would be:
\begin{equation}
\mathcal{L}(\alpha w) = \frac{\alpha w^\top A (\alpha w)}{\alpha w^\top B (\alpha w)} = \frac{\alpha^2 (w^\top A w)}{\alpha^2 (w^\top B w)}
\label{eq:13}
\end{equation}

The $a^2$ gets offset, we could get $\mathcal{L}(\alpha w) = \frac{w^\top A w}{w^\top B w} = \mathcal{L}(w)$, which denotes that \textbf{Deep LDA is a Homogeneous function of degree $0$}, this is the first finding.

Tracing back the theoretical setup we did in section 3.1.1. We prove $\frac{\partial w_i}{\partial u_i^{(k)}} = \frac{w_i}{u_i^{(k)}}$(simply take partial derivative) and $\frac{du_i^{(k)}}{dt} = - \frac{\partial \mathcal{L}}{\partial w_i} \left( \frac{w_i}{u_i^{(k)}} \right)$. So the total derivative $\frac{dw_i}{dt}$ could convert to an expression about the loss function in eq.\eqref{eq:14}.

\begin{equation}
\frac{dw_i}{dt} = \sum_{k=1}^L \frac{\partial w_i}{\partial u_i^{(k)}} \frac{du_i^{(k)}}{dt}=\sum_{k=1}^L \left( \frac{w_i}{u_i^{(k)}} \right) \left[ - \frac{\partial \mathcal{L}}{\partial w_i} \left( \frac{w_i}{u_i^{(k)}} \right) \right] = - \frac{\partial \mathcal{L}}{\partial w_i} \sum_{k=1}^L \frac{w_i^2}{\left( u_i^{(k)} \right)^2}
\label{eq:14}
\end{equation}

Because $u_i^{(k)}(t)$ get the result is $w_i(t)^{\frac{1}{L}}$ under balance initialization,we could replace the denominator. The eq.\eqref{eq:15} denotes a nonlinear gradient flow naturally induces sparsity-like behavior, which explains multiplicative update in contrast to standard linear ones.

\begin{equation}
\frac{dw_i}{dt} = - \frac{\partial \mathcal{L}}{\partial w_i} \sum_{k=1}^L \frac{w_i^2}{w_i^{\frac{2}{L}}}= - L \cdot w_i^{2 - \frac{2}{L}} \frac{\partial \mathcal{L}}{\partial w_i}
\label{eq:15}
\end{equation}

Since we prove Scale Invariance in eq.\eqref{eq:13},and its have natural property in eq.\eqref{eq:16}.(see \textbf{Appendix A} or the book Matrix Computations\cite{Matrix_Computations} for the complete proof, in geometrical aspect to understand, only when it orthogonal to the direction of weight could cause it not effected by scaling)

\begin{equation}
w^\top \nabla_w \mathcal{L} = \sum w_i \frac{\partial \mathcal{L}}{\partial w_i} = 0
\label{eq:16}
\end{equation}

And we further multiply the both sides of the eq.\eqref{eq:15} with $w_i^{\frac{2}{L} - 1}$.

\begin{equation}
w_i^{\frac{2}{L} - 1} \frac{dw_i}{dt} = - L \cdot w_i^{\left(2 - \frac{2}{L}\right) + \left(\frac{2}{L} - 1\right)} \frac{\partial \mathcal{L}}{\partial w_i} w_i^{\frac{2}{L} - 1} = - L \cdot w_i \frac{\partial \mathcal{L}}{\partial w_i}
\label{eq:17}
\end{equation}

Taking the integral again to both sides of eq.\eqref{eq:17}:

\begin{equation}
\frac{d}{dt} \left( w_i^{\frac{2}{L}} \right) = - 2 w_i \frac{\partial \mathcal{L}}{\partial w_i}
\label{eq:18}
\end{equation}

\begin{equation}
\frac{d}{dt} \sum_{i=1}^d w_i^{\frac{2}{L}} = - 2 \sum_{i=1}^d w_i \frac{\partial \mathcal{L}}{\partial w_i} = - 2 w^\top \nabla_w \mathcal{L}(w)
\label{eq:19}
\end{equation}

Because Scale Invariance in eq.\eqref{eq:16},we could get an expression for the sum up of weights in all dimensions is get balanced and bound with a constant $C$.
\begin{equation}
\sum_{i=1}^d w_i(t)^{\frac{2}{L}} = C
\label{eq:20}
\end{equation}

To other words, the whole training process, \textbf{the Deep LDA is bounded by the $|w(t)|{2/L}$ } as the expression show beneath:
\begin{equation}
|w(t)|{2/L}^{2/L} = |w(0)|_{2/L}^{2/L} \quad \forall t \ge 0
\label{eq:21}
\end{equation}

\section{Experiment}

\subsection{Simulation Test}
Our first experiment settled a test in DLNs implemented with the basic Numpy package, testing with hyper-parameters( $\eta=0.005$ and $100000$ epochs). The test is mainly for comparing the dynamic differences between the networks with different numbers of layers $L$, including $L=[1, 2, 5, 10, 20]$, since the depth $L$ is highly dependent to the effects as we discuss in \eqref{eq:20}. For the experimental setup, we defined a simplified feature space with a dimension of $d=5$, and synthesized the intra-class and inter-class scatter matrices ($S_w$ and $S_b$) as random positive-definite matrices from a uniform distribution $\mathcal{U}(0.4, 0.6)$.

\begin{figure}[h]
\centering
\includegraphics[width=1\linewidth]{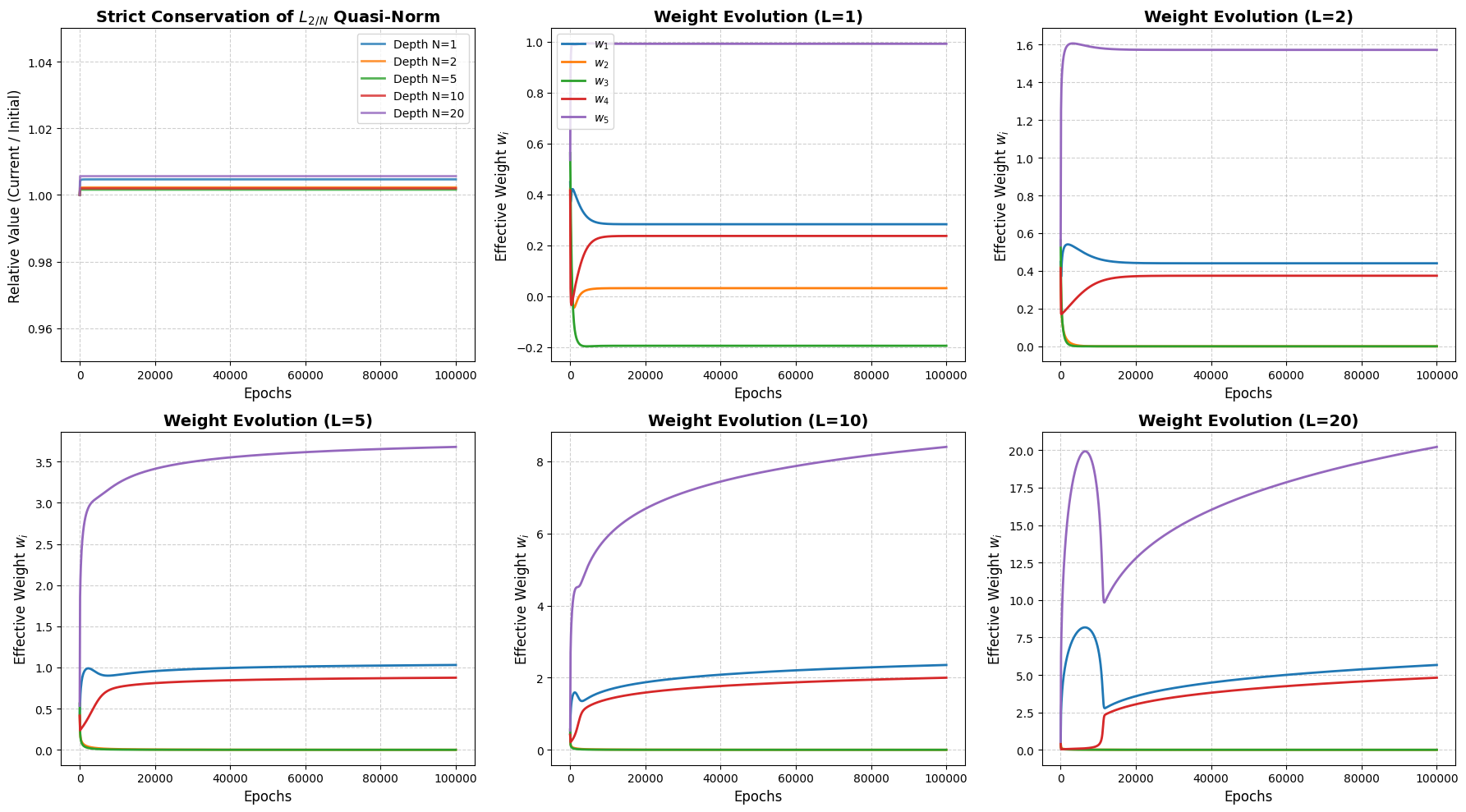}
\caption{The Simulated Result for DeepLDA in DLNs.}
\label{fig:2}
\end{figure}

In the top-left subplot of fig.\ref{fig:2}, we find five horizontal plain lines,which validate our conclusion in eq.\eqref{eq:9}, no matter how many layers a DLN gets, the implicit regularization remains unchanged.

About the second subplot to the last one, the ones about weight evolution, it clearly denotes with an increasing number of layers. With a smaller number of layers, the weak features get eliminated faster than the DLNs, and slower for the strong features converge. The weak features are selected out faster under the expectation due to our conclusion in eq.\eqref{eq:20}. The phenomenon that strong features get fluctuating is due to the Edge of Stability caused by our stable learning rate,since in previous conduct we were all using $dt$ rather than a constant learning rate.

\subsection{Real-World Data Validation}

To further validate our method that not only establishes in ideal simulated data, we also test the same step in three classical real-world dataset. The complexity of the dataset is escalating with dimensions to test the impact it has on the theory. Including the Iris dataset(dim=4,3 class,150 samples)\cite{FDA}, UCI wine dataset(dim=13,3 class,178 samples)\cite{Wine} and Wisconsin Breast Cancer Dataset (dim=30, 2 class, 569 samples)\cite{Cancer}. 

The reasons for utilizing the Iris Dataset are that Fisher's paper that suggest LDA in first time, using it as an example in his experiment, ensure its good separability in class, and the final result is explicit to check. As for the UCI wine dataset, we using it as a supplement since its similar data structure compare to the Iris Dataset(both get three class),which could control the number of classification while still observe the influence of larger dimensions. The Wisconsin Breast Cancer Dataset is for settling extreme conditions to see how the sparsity get involve.Since the number of class $c$ smaller than the input dimension $dim$,we adding a minimal value to $S_b$ to ensure its positive-definite. 

Maximum deviation is defined to measure relatively value change compare its original value, the detailed expression is:
\begin{equation}
\Delta_{\max} = \max_{t} \left| \frac{\|w(t)\|_{2/L}^{2/L}}{\|w(0)\|_{2/L}^{2/L}} - 1 \right|
\label{eq:22}
\end{equation}

Table.\ref{tab:1} reports the maximum deviation below 0.6\% verify our derivation,which not only prove the conservation to some extent in eq.\eqref{eq:20} but also prove the implicit bias of this gradient dynamic plays a dominant role in the process.

\begin{table}[ht]
\caption{Quasi-Norm Conservation: Maximum Deviation}
\centering
\begin{tabular}{@{}llllll@{}}
\toprule
Dataset & $L=1$ & $L=2$ & $L=5$ & $L=10$ & $L=20$ \\
\midrule
Simulated ($d\!=\!5$)      & 0.466\% & 0.226\% & 0.157\% & 0.194\% & 0.562\% \\
Iris ($d\!=\!4$)            & 0.076\% & 0.031\% & 0.020\% & 0.017\% & 0.018\% \\
Wine ($d\!=\!13$)           & 0.239\% & 0.106\% & 0.071\% & 0.083\% & 0.209\% \\
Breast Cancer ($d\!=\!30$)  & 0.038\% & 0.016\% & 0.010\% & 0.009\% & 0.009\% \\
\bottomrule
\end{tabular}
\label{tab:1}
\end{table}

Across all four datasets, we measure sparsity by the normalized criterion $|w_i / w_{\max}| < 0.01$, which identifies features whose final weight is less than $1\%$ of the dominant feature, effectively eliminated from the discriminant direction. 

The primary sparsity transition occurs at $L=2$, with a minor refinement at $L=5$ for the highest-dimensional dataset, after which the sparsity pattern stabilizes.

Prior works have established that MSE induces nuclear norm minimization \cite{Square1} and logistic loss converges in the max-margin direction \cite{Cross1}. Our result reveals that scale-invariant objectives constitute a distinct class with a fundamentally different implicit bias: strict quasi-norm conservation rather than directional convergence.

\begin{figure}[h]
\centering
\includegraphics[width=1\linewidth]{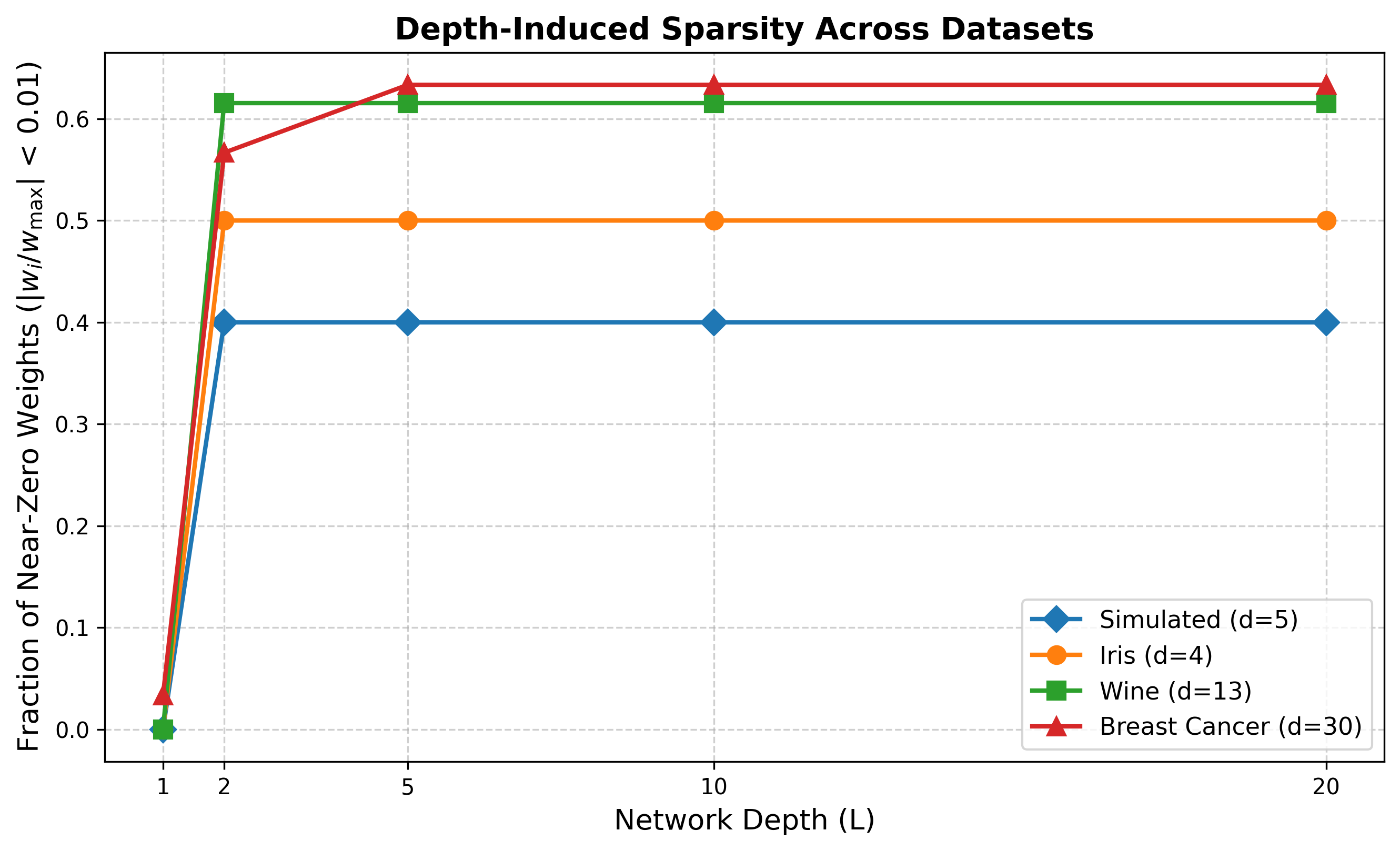}
\caption{Depth-induced feature sparsity across datasets}
\label{fig:3}
\end{figure}

\section{Conclusion}
In conclusion, we provide a theoretical perspective on the implicit bias induced by Deep Linear Discriminant Analysis (Deep LDA) under a diagonal linear network (DLN) framework. Crucially, this conservation stem from two key ingredients: (1) the multiplicative parameterization induced by depth, and (2) the scale-invariance property of the Rayleigh quotient objective. Together, these yield a strict geometric constraint on the optimization trajectory. The experiment finds that deeper architectures amplify multiplicative penalties on weak features, promoting sparsity-like behavior in the effective weights.

Our current findings are fundamentally constrained by the simplified diagonal linear architecture without non-linear activations, and only been testing in small-scale of data samples, need further validations for real-world data. Without balanced initialization, the conservation in \eqref{eq:9} no longer holds exactly, and the layer weights may diverge, breaking the subsequent quasi-norm conservation. Future works will extend this continuous-time framework beyond diagonal networks to incorporate non-linear activation functions. Additionally, investigating how stochastic gradient descent impacts this strict quasi-norm conservation in Deep LDA remains a critical open direction.  The  strict conservation law about quasi-norm index  $2/L$ suggests a guideline for model architecture design, allowing them to reduce explicit regularization term which defined manually.

\appendix
\section{Proof of the Orthogonality Lemma}\label{secA1}
In this section, we provide the rigorous mathematical analysis for the gradient of the Deep LDA objective and prove the orthogonality property $w^\top \nabla_w \mathcal{L}(w) = 0$ referenced in Section 3.2.

Let the generalized Rayleigh Quotient be defined as:
\begin{equation}
\mathcal{L}(w) = \frac{N(w)}{D(w)} = \frac{w^\top S_w w}{w^\top S_b w}
\end{equation}
where the numerator is $N(w) = w^\top S_w w$ and the denominator is $D(w) = w^\top S_b w$.

\subsection{Gradient of the Rayleigh Quotient}
Because both the intra-class scatter matrix $S_w$ and the inter-class scatter matrix $S_b$ are symmetric real matrices, the derivative of their associated quadratic forms with respect to the vector $w$ can be straightforwardly evaluated as:
\begin{equation}
\nabla_w N(w) = 2 S_w w, \quad \nabla_w D(w) = 2 S_b w
\end{equation}

Applying the standard quotient rule for differentiation, $\nabla \left( \frac{N}{D} \right) = \frac{D \nabla N - N \nabla D}{D^2}$, we obtain the gradient of $\mathcal{L}(w)$:
\begin{equation}
\nabla_w \mathcal{L}(w) = \frac{(w^\top S_b w)(2 S_w w) - (w^\top S_w w)(2 S_b w)}{(w^\top S_b w)^2}
\end{equation}

By factoring out $\frac{2}{w^\top S_b w}$, we can rewrite the gradient elegantly in terms of $\mathcal{L}(w)$ itself:
\begin{equation}
\nabla_w \mathcal{L}(w) = \frac{2}{w^\top S_b w} \left( S_w w - \left(\frac{w^\top S_w w}{w^\top S_b w}\right) S_b w \right) = \frac{2}{w^\top S_b w} \big( S_w w - \mathcal{L}(w) S_b w \big)
\label{eq:appendix_grad}
\end{equation}

\subsection{Scale Invariance Property}
Using the derived gradient in Eq. \eqref{eq:appendix_grad}, we can now calculate the inner product between the weight vector $w$ and its gradient $\nabla_w \mathcal{L}(w)$:
\begin{equation}
w^\top \nabla_w \mathcal{L}(w) = w^\top \left[ \frac{2}{w^\top S_b w} \big( S_w w - \mathcal{L}(w) S_b w \big) \right]
\end{equation}

Since $\frac{2}{w^\top S_b w}$ is a scalar, we distribute $w^\top$ into the parenthesis:
\begin{equation}
w^\top \nabla_w \mathcal{L}(w) = \frac{2}{w^\top S_b w} \Big( w^\top S_w w - \mathcal{L}(w) w^\top S_b w \Big)
\end{equation}

Substituting the original definition of $\mathcal{L}(w) = \frac{w^\top S_w w}{w^\top S_b w}$ back into the second term:
\begin{equation}
w^\top \nabla_w \mathcal{L}(w) = \frac{2}{w^\top S_b w} \left( w^\top S_w w - \left( \frac{w^\top S_w w}{w^\top S_b w} \right) w^\top S_b w \right)
\end{equation}

The scalar $w^\top S_b w$ cancels out perfectly, leaving:
\begin{equation}
w^\top \nabla_w \mathcal{L}(w) = \frac{2}{w^\top S_b w} \Big( w^\top S_w w - w^\top S_w w \Big) = 0
\end{equation}
This completes the proof that the gradient of the scale-invariant Deep LDA objective is always orthogonal to the weight vector itself.

\begin{credits}
\subsubsection{\ackname} 
The authors declare that they have no known competing financial interests or personal relationships that could have appeared to influence the work reported in this paper.

\subsubsection{\discintname}
\textbf{Data Availability Statement:} The data were generated using \texttt{numpy.random.randn} with a fixed seed (8086) for reproducibility,other three experimented data testing with package sklearn. Upon publication, the experiment will be available at \url{https://www.kaggle.com/code/spike8086/implicit-bias-in-deep-lda}. \\
\textbf{Funding Information:} Not Applicable. \\
\textbf{Author Contribution:} Not Applicable. \\
\textbf{Research Involving Human or Animals / Informed Consent:} Not Applicable.
\end{credits}

\bibliographystyle{splncs04}
\bibliography{sn-bibliography}

\end{document}